\pdfoutput=1

\documentclass[11pt]{article}

\usepackage[final]{acl}
\usepackage{geometry}
\usepackage{longtable}
\usepackage{multicol}
\usepackage{multirow}
\usepackage{times}
\usepackage{latexsym}
\usepackage{comment}
\usepackage[T1]{fontenc}

\usepackage[utf8]{inputenc}
\usepackage{hyperref}
\usepackage{microtype}

\usepackage{inconsolata}

\usepackage{graphicx}

\title{Overview of MWE history, challenges, and horizons: standing at the 20th anniversary of the MWE workshop series via MWE-UD2024}

\author{Lifeng Han$^1$, 
Kilian Evang$^2$, \\
\textbf{Archna Bhatia$^3$,
Gosse Bouma$^4$,
A. Seza Doğru\"oz$^5$,}
\\
\textbf{Marcos Garcia$^6$,
Voula Giouli$^7$,
Joakim Nivre$^8$,
Alexandre Rademacher$^9$}\\
$^1$LIACS \& LUMC, Leiden University; University of Manchester; 	\\
$^2$Heinrich Heine University Düsseldorf;	
$^3$Institute for Human and Machine Cognition, USA;	\\$^4$Groningen University;	$^5$Ghent University; $^6$University of Santiago de Compostela;\\	$^7$Institute for Language \& Speech Processing, ATHENA RC, Greece;	\\
$^8$Uppsala University and Research Institutes of Sweden, Sweden;	$^9$IBM Research, Brazil}

\begin{document}
\maketitle
\begin{abstract}

Starting in 2003 when the first MWE workshop was held with ACL in Sapporo, Japan, this year, the joint workshop of MWE-UD co-located with the LREC-COLING 2024 conference marked the 20th anniversary of MWE workshop events over the past nearly two decades.
Standing at this milestone, we look back to this workshop series and summarise the research topics and methodologies researchers have carried out over the years. We also discuss the current challenges that we are facing and the broader impacts/synergies of MWE research within the CL and NLP fields. Finally, we give future research perspectives. We hope this position paper can help researchers, students, and industrial practitioners interested in MWE get a brief but easy understanding of its history, current, and possible future.
 \\ \newline Keywords: {MWE, UD, history, challenges, impacts, horizon} 

\end{abstract}

\section{Introduction}


Multiword Expressions (MWEs) are not only a linguistic phenomenon but also known as a pain in the neck of natural language processing (NLP) \cite{2002Sag_10.1007/3-540-45715-1_1}. 
The workshop series
dedicated to Multiword Expressions (MWEs) research is embracing its 20th anniversary via the MWE-UD 2024 workshop since the first one (MWE 2003) was held in 2003, co-located with ACL \cite{ws-2003-acl-2003-multiword}. 
The MWE workshop series is organised by the MWE section of SIGLEX -- Special Interest Group on Lexicon -- of the Association for Computational Linguistics \footnote{\url{https://multiword.org/}}. MWE workshops are dedicated to promoting scientific research on MWEs in computational linguistics (CL) and NLP settings.
The MWE workshop has always been co-located with major CL and NLP conferences such as the Coling, *ACL, and LREC series, drawing attention from a broad range of researchers. The previous 19 proceedings volumes are available at \url{https://aclanthology.org/venues/mwe/}.

There have been funded international and local projects that support the organisation of this event, such as the PARSEME research network (PARSing and Multi-word Expressions)\footnote{\url{http://parseme.eu/}} stemming from the eponymous COST Action (2013--2017), which helped organise resource creation, meetings, and training. One of the major outcomes of PARSEME is the PARSEME corpus, annotated in a multidisciplinary setting by native speakers across different countries covering 26 languages \cite{savary-etal-2023-parseme}. There have been 3 
shared tasks on MWE identification and discovery based on the PARSEME corpus in 2017 \cite{savary-etal-2017-parseme}, 2018 \cite{ramisch-etal-2018-edition}, and 2020 \cite{ramisch-etal-2020-edition}. 
From 2017 on, there has been the Universal Dependencies (UD) framework formed and funded by the EU to promote cross-lingual consistent treebank annotations \cite{ud2017proceedings}, which involves MWE annotations and resource creation. 

\section{Looking Back to History}
\label{sec:history}

Looking back to the MWE workshop historical events, there have been some designed and/or summarised themes across the years, in addition to non-specified ones \cite{ws-2004-multiword,ws-2013-multiword,ws-2014-multiword,ws-2015-multiword,ws-2016-multiword,ws-2017-multiword,mwe-2021-multiword,mwe-2022-multiword}, such as:
\begin{itemize}
    \item MWE Analysis, Acquisition and Treatment at MWE 2003 \cite{ws-2003-acl-2003-multiword}
    \item MWE in a Multilingual Context, MWE 2006 \cite{ws-2006-multi}
    \item Identifying and Exploring Underlying Properties at MWE 2006 \cite{ws-2006-multiword_underlying}
    \item MWEs in broader perspectives at MWE 2007 \cite{ws-2007-broader}
    \item MWE identification, interpretation, disambiguation and applications at MWE 2009 \cite{ws-2009-multiword}
    \item from Theory to Applications at MWE 2010 \cite{ws-2010-2010-multiword}
    \item from Parsing and Generation to the Real World at MWE 2011 \cite{ws-2011-multiword}
    \item Linguistic Annotation, MWEs and Constructions at MWE 2018 \cite{ws-2018-joint-linguistic} (Joint LAW-MWE-CxG-2018)
    \item MWEs and WordNet at MWE 2019 \cite{ws-2019-joint-multiword} (Joint MWE-WN 2019)
    \item MWEs and Electronic Lexicons at MWE 2020 \cite{mwe-2020-joint}
    \item MWEs in Knowledge-Intense Clinical Domain at MWE 2023 \cite{mwe-2023-multiword} 
\end{itemize}

Some details of the three organised shared tasks from PARSEME (data available at \url{https://gitlab.com/parseme/}) on verbal MWEs (vMWEs) are below \cite{savary-etal-2017-parseme,ramisch-etal-2018-edition,ramisch-etal-2020-edition}.
\begin{itemize}
    \item Task v1.0: 18 languages and  vMWEs only. 
\textbf{Balto-Slavic}: Bulgarian (BG), 
Czech (CS), Lithuanian (LT), Polish (PL) and Slovene (SL); 
\textbf{Germanic}: 
German (DE), Swedish (SV); 
\textbf{Romance}: French (FR), Italian (IT), Romanian (RO),
Spanish (ES) and Brazilian Portuguese (PT); 
and
\textbf{others}: Farsi (FA), Greek (EL), Hebrew (HE),
Hungarian (HU), Maltese (MT) and Turkish (TR).
Most of them are European languages, except for 4 non-Indo-European ones, i.e. HE, HU, MT, and TR.

    \item Task v1.1: 19 languages on vMWEs. 
    \item Task v1.2: 14 languages on vMWEs, focusing on identifying \textbf{unseen} vMWEs. New languages or substantially increased size in comparison to 1.0/1.1: Irish (GA), Swedish (SV), and Chinese (ZH)
 ``7 teams who submitted 9 system results''
\end{itemize}
The evaluation settings for the MWE-shared tasks have been improved and standardised over the years. For instance, in MWE-2020-v1.2, the evaluations include both \textit{global} and \textit{token-based} precision, recall, and F-score per language via macro-averages. There are also evaluations on specific vMWE phenomena including continuous and discontinuous ones, and categories, e.g., different vMWE types including VID, IRV, LVC.full, etc. 
Systems were also ranked separately by closed track and open track, of which the closed-track is restricted to only using the provided corpus as resources.
The top-performing system in the closed track used BiLSTM-CRF structure, while the top-performing systems in the open track used fine-tuned neural LMs in either multilingual or mono-lingual settings. 
The external resources used included 1) dataset-wise: vMWE lexicons, doom datasets, FrWac corpus, Wiktionary, CoNLL corpus; 2) model-wise mBERT, monolingual BERT, XLM-RoBERTa. \cite{ramisch-etal-2020-edition}

At a similar time frame since the first vMWE identification shared task was launched, there has been the PMWE book series dedicated on the topic of ``Phraseology and Multiword Expressions'' published by Language Science Press with six volumes so far \footnote{\url{https://langsci-press.org/catalog/series/pmwe}}:

\begin{itemize}
    \item (Volume 1) ``Multiword expressions: Insights from a multi-lingual perspective'' \cite{sailer2018multiword}
    \item (Volume 2) ``Multiword expressions at length and in depth: Extended papers from the MWE 2017 workshop'' \cite{markantonatou2018multiword}
    \item (Volume 3) ``Representation and parsing of multiword expressions: Current trends'' \cite{parmentier2019representation}
    \item (Volume 4) ``The role of constituents in multiword expressions: An interdisciplinary, cross-lingual perspective'' \cite{smolka2020role}
    \item (Volume 5) ``Formulaic language: Theories and methods'' \cite{trklja2021formulaic}
    \item Forthcoming ``Multiword expressions in lexical resources: Linguistic, lexicographic, and computational perspectives''
\end{itemize}

MWEs are also the focus of book chapters in popular NLP handbooks
\cite{baldwin2010multiword}. 
\newcite{ramisch2015multiword} wrote a book about MWE acquisitions using the MWEToolkit \cite{ramisch-etal-2010-mwetoolkit}. 
Recent doctoral theses on MWE research 
explored the strategies for improving MWE translation by augmented bilingual MWE lexicons in Transformer structures and character decomposition when it comes to pictogram languages such as Chinese \cite{han2022investigation}.

\section{Synergies of MWEs}


The research on MWEs has a broad impact and synergy with other related CL and NLP fields. This has been reflected by editions of the MWE workshop that were held as joint workshops with other communities, \emph{viz.} the Joint Workshop on Linguistic Annotation, MWEs and Constructions (LAW-MWE-CxG) in 2018, the Joint Workshop on MWEs and WordNet in 2019, the Joint Workshop on MWEs and Electronic Lexicons in 2020, and the Joint Workshop on MWEs and Universal Dependencies in 2024 (cf. Section~\ref{sec:history}). 
The joint track we introduced in MWE 2023 with Clinical-NLP also opens a window to tackling the scientific difficulty of addressing domain-specific knowledge.
Looking back to the vMWE identification and discovery shared task, the methodologies researchers proposed overlap with the areas of named entity recognition (NER), text mining, semantic compositionality detection, domain-language adaptation, e.g., social media data, etc.
The research of MWEs also influences some traditional NLP tasks including part-of-speech (POS) tagging, parsing, machine translation, and summarisation.

\section{Challenges and Opportunities}

The vMWE shared tasks have shown that 
unseen vMWEs present greater challenges than other vMWEs, with F1-scores from top performing systems in the 20-30\% range compared with 50-70\% at MWE-based evaluations with the corpus setting from MWE-v1.2 \cite{ramisch-etal-2020-edition}. 

MWEs, especially idiomatic and metaphorical expressions, still cause pain in many NLP tasks, such as machine translation where the state-of-the-art NMT systems fail to produce meaning-equivalent translations, not to mention meaning-equivalent idiomatic expressions \cite{han2022investigation}.

There is a need for new data sets and language resource creation regarding non-verbal MWEs, such as nominal compounds, collocations, idioms, metaphors, etc., to facilitate non-verbal MWE research and experimental testing of current NLP models.

Opportunities for MWE research in the new era with \textbf{emerging} study fields include MWE-detection and analysis in online forums, abusive language detection for social good, language learning and assessments, text simplification and domain-adaptation, LLM hallucination \cite{li2023largeLM_biomedReadability}, etc. 

With the popularity of \textbf{LLMs}, their learning interpretation such as how neural layers perform on the detection of idiomatic phrases becomes an urgent need. For example, \newcite{nedumpozhimana2024topicProb} investigated the relationship between topic encoding and idiomatic verb-noun phrase detection between the BERT and RoBERTa layers.


\section{Acknowledging the Giant Shoulders}
We sincerely thank the historical MWE WS organisers, MWE standing committees, funding supports/projects that helped MWE workshop series keep on going, and the historical \textbf{handbooks} the committees prepared for us. 

\section{Language Resources}
For the convenience of readers, there is the official PARSEME corpus v1.3 \cite{savary-etal-2023-parseme} available at \url{https://parsemefr.lis-lab.fr/parseme-st-guidelines/1.3/?page=home}.  
There are also other MWE corpora available created by researchers separately or as extended resources from PARSEME.

Multilingual MWEs resources derived from PARSEME corpus, such as:
\begin{itemize}
    \item AlphaMWE: from PARSEME English to MT+PE generated Polish, German, Chinese \cite{han-etal-2020-alphamwe}, and Arabic \cite{hadj-mohamed-etal-2023-alphamwe}.
\end{itemize}

Multilingual MWEs resources independent from PARSEME, such as:
\begin{itemize}
    \item MultiMWE: English-German and English-Chinese bilingual MWE lexicons extracted automatically from 5 million WMT parallel sentences respectively \cite{han-etal-2020-multimwe}
\end{itemize}

More resources created by authors can be found in this year's MWE-UD2024 proceeding \cite{mwe-2024-joint} covering the languages:
Arabic, Hindi, Old Egyptian, Vedic, Northern Kurdish, Slovene, Dutch, Bavarian, South Asian languages, Turkic languages, Gujarati, Saraiki, Swedish, and more. 


\nocite{*}
\section{Bibliographical References}\label{sec:reference}

\bibliography{custom}


\appendix

\section{Recent MWE Events}
The Table of Content for MWE-2023 and MWE-UD2024 can be found at \url{https://multiword.org/mwe2023/} and \url{https://multiword.org/mweud2024/}. They are detailed below.

\subsection{MWE-2023  with ClinicalNLP}
The MWE-2023 WS held with EACL-2023 features the synergy with the 5th Clinical-NLP WS which was held in conjunction with ACL-2023. There are two set keynote speakers: 1) on MWE as tradition by Prof. Leo Wanner ``Lexical collocations: Explored a lot, still a lot
more to explore'', 2) on MWE and Clinical-NLP synergy track by Prof. Goran Nenadic and Dr. Asma Ben Abacha ``MWEs in ClinicalNLP and Healthcare Text Analytics''. 
The workshop also features a panel discussion on ``Multiword Expressions in Knowledge-intensive Domains: Clinical Text as a Case Study''.
The Proceedings include 15 papers covering two tracks: the MWE traditional track and the MWE+ClinicalNLP track, with the best paper award to the joint track (\url{https://aclanthology.org/volumes/2023.mwe-1/}) \cite{mwe-2023-multiword}.

\subsection{MWE-UD 2024 Accepted Papers}
MWE-UD2024 Proceedings feature 8 UD track papers, 9 MWE track papers, and 9 MWE+UD papers (out to 52 submissions), at
\url{https://aclanthology.org/volumes/2024.mwe-1/}. 
In addition, there are 8 non-archival presentations, with the paper list and presentation posters/slides publicly available (at: \url{https://multiword.org/mweud2024/}) \cite{mwe-2024-joint}.

\subsection{MWE-2025 at NAACL}
The next edition of the 21st MWE WS will be held at NAACL-2025, location:  Albuquerque, New Mexico, U.S.A;
Date of the Workshop: May 3/4, 2025 \url{https://multiword.org/mwe2025/}.

\end{document}